\documentclass[letterpaper, 10 pt, conference]{ieeeconf}  
\let\labelindent\relax
\usepackage{enumitem}

\IEEEoverridecommandlockouts                              

\usepackage{amsmath,amsfonts}
\usepackage{booktabs}
\usepackage{graphicx}
\usepackage{xcolor}
\usepackage{multirow}
\usepackage{hyperref}       
\usepackage{cleveref}
\usepackage{siunitx}   
\usepackage{cuted} 
\usepackage{arydshln}
\usepackage{float} 
\usepackage{enumitem} 

\title{\LARGE \bf
GP-enhanced Autonomous Drifting Framework using ADMM-based iLQR
}

\author{Yangyang Xie$^{1}$, Cheng Hu$^{1}$, Nicolas Baumann$^{2}$, Edoardo Ghignone$^{2}$, Michele Magno$^{2}$, and Lei Xie$^{1,3}$
\thanks{$^{1}$Yangyang Xie, Cheng Hu and Lei Xie are with State Key Laboratory of Industrial Control Technology, Zhejiang University, Hangzhou, 310027, Zhejiang, China
        {\tt\small \{22332009|22032081\}@zju.edu.cn, leix@iipc.zju.edu.cn}. }%
\thanks{$^{2}$Nicolas Baumann, Edoardo Ghignone, and Michele Magno are associated with the Center for Project-Based Learning, D-ITET, ETH Zurich {\tt\small \{nibauman|eghignone\}@pbl.ee.ethz.ch, michele.magno@pbl.ee.ethz.ch}.}%
\thanks{$^{3}$The corresponding author of this paper}
\thanks{\emph{(Yangyang Xie and Cheng Hu contributed equally to this work.)}}
}

\begin{document}

\maketitle
\thispagestyle{empty}
\pagestyle{empty}

\begin{figure*}
    \centering
    \includegraphics[width=1\textwidth]{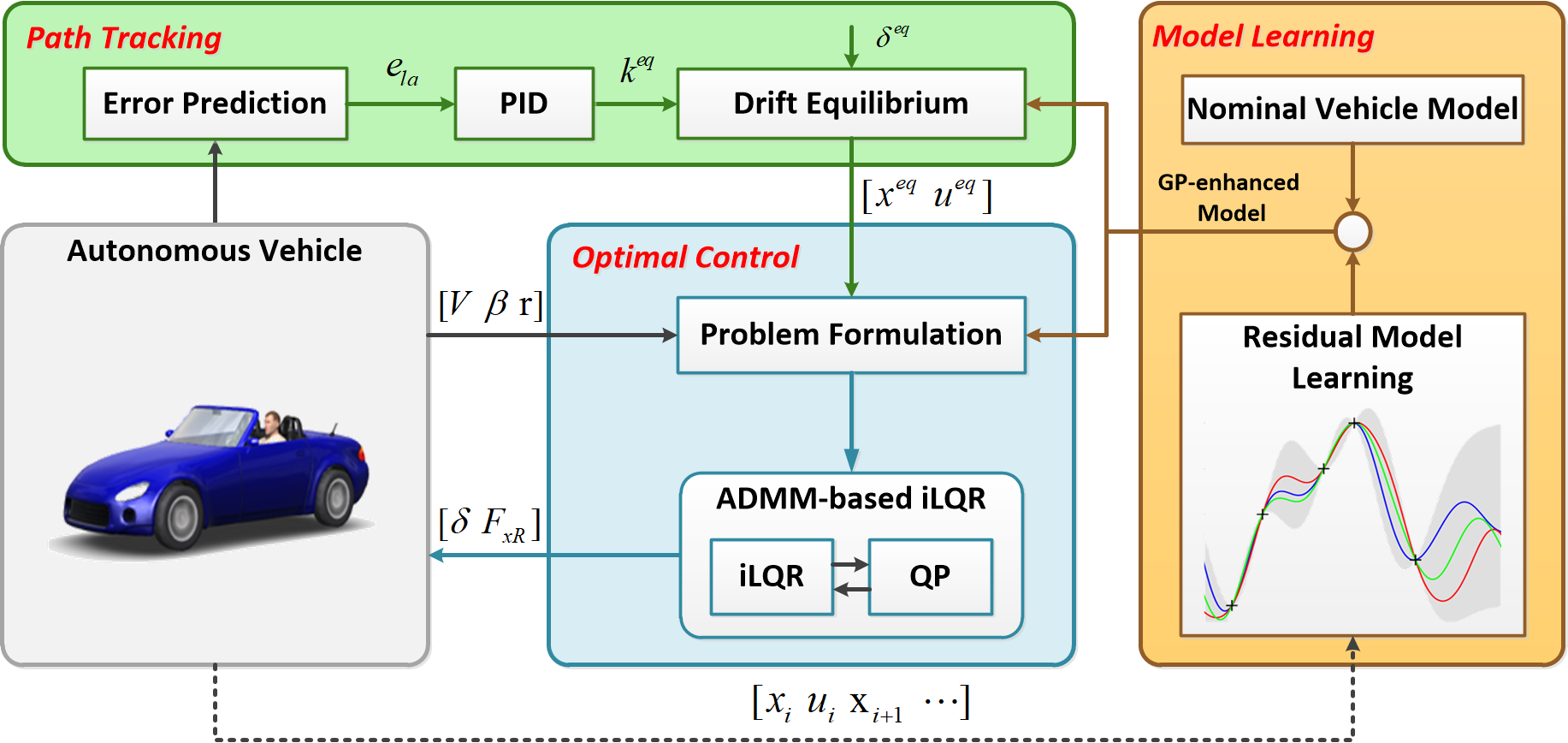}
    \caption{The framework integrates three key components: Path Tracking, Optimal Control, and Model Learning. The Path Tracking layer generates the desired states based on predicted errors. The Optimal Control layer leverages the iterative Linear Quadratic Regulator (iLQR) to efficiently solve the nonlinear constrained control problem in real-time, with the Alternating Direction Method of Multipliers (ADMM) aiding in problem decomposing. To account for model inaccuracies and environmental uncertainties, the Model Learning layer employs Gaussian Process (GP) regression to compensate for dynamic residuals.}
    \label{fig:Overall system architecture}
\end{figure*}

\begin{abstract}

Autonomous drifting is a complex challenge due to the highly nonlinear dynamics and the need for precise real-time control, especially in uncertain environments. To address these limitations, this paper presents a hierarchical control framework for autonomous vehicles drifting along general paths, primarily focusing on addressing model inaccuracies and mitigating computational challenges in real-time control. The framework integrates Gaussian Process (GP) regression with an Alternating Direction Method of Multipliers (ADMM)-based iterative Linear Quadratic Regulator (iLQR). GP regression effectively compensates for model residuals, improving accuracy in dynamic conditions. ADMM-based iLQR not only combines the rapid trajectory optimization of iLQR but also utilizes ADMM's strength in decomposing the problem into simpler sub-problems. Simulation results demonstrate the effectiveness of the proposed framework, with significant improvements in both drift trajectory tracking and computational efficiency. Our approach resulted in a 38$\%$ reduction in RMSE lateral error and achieved an average computation time that is 75$\%$ lower than that of the Interior Point OPTimizer (IPOPT).

\end{abstract}

\section{INTRODUCTION}
Autonomous driving technologies have made significant strides in recent years, yet performing high-dynamic maneuvers such as drifting remains a complex challenge. Drifting, characterized by the intentional over-steering of a vehicle to induce loss of rear-wheel traction while maintaining control through the front wheels, has garnered sustained attention in both academic and industrial research due to its unique dynamics and control challenges \cite{lenzo2024autonomous} \cite{jia2024novel}. This maneuver, while being an essential aspect of advanced driving, also presents considerable difficulties in terms of vehicle control, particularly when considering the highly nonlinear tire-road interactions and fast-changing dynamics involved.

Most existing drift control methods rely heavily on accurate dynamic models to stabilize the vehicle during these high-speed maneuvers. Traditional approaches, such as those in  \cite{goh2016simultaneous} and \cite{goh2020toward}, explore error dynamic models and nonlinear model inversion to achieve the desired state trajectories for effective drift control. Similarly, optimization-based techniques like Model Predictive Control (MPC) have been widely adopted to manage the vehicle’s dynamic states around drift equilibrium points \cite{hu2022combined} \cite{hu2024novel} \cite{bellegarda2021dynamic} \cite{goh2024beyond}. However, while these methods have proven effective under well-modeled conditions, they remain highly dependent on the accuracy of the underlying vehicle dynamics model. Even small deviations in model parameters, such as a $2\%$ change in road friction, can result in significant performance degradation \cite{laurense2017path}. This reliance on precise models makes traditional methods vulnerable in uncertain or rapidly changing environments.

To address these limitations, machine learning techniques have been integrated, offering adaptability to unmodeled dynamics and improved model accuracy \cite{zhou2022learning} \cite{djeumou2023autonomous}. Among machine learning techniques, Gaussian Processes (GP) have shown particular promise. Studies such as \cite{hewing2019cautious} \cite{kabzan2019learning} and \cite{zhou2024learning} demonstrated GP’s effectiveness in capturing dynamic uncertainties and improving control performance.

However, the integration of GP into real-time control frameworks introduces significant computational challenges due to its nonparametric nature \cite{kabzan2019learning}. This creates a trade-off between model accuracy and computational efficiency, which becomes particularly problematic when combined with Nonlinear Model Predictive Control (NMPC), where real-time performance is essential.

To manage these computational demands and ensure effective constraint handling, the combination of the iterative Linear Quadratic Regulator (iLQR) \cite{li2004iterative} and the Alternating Direction Method of Multipliers (ADMM) \cite{boyd2011distributed} has emerged as a promising solution. iLQR is known for its efficiency in optimizing control trajectories for nonlinear systems, and recent studies have extended its application to handle more complex, constrained optimization tasks \cite{tassa2014control} \cite{howell2019altro} \cite{chen2019autonomous}. To further enhance iLQR’s ability to manage physical constraints, ADMM is employed as a complementary method. ADMM restructures the constrained optimization problem into smaller sub-problems, solving them sequentially while respecting the system's physical limitations. ADMM has demonstrated success in various optimal control applications \cite{nguyen2024tinympc} \cite{ma2022alternating} \cite{sindhwani2017sequential}, making it a natural fit alongside iLQR without sacrificing real-time performance.

This paper presents a novel hierarchical control framework (Fig.\ref{fig:Overall system architecture}) for autonomous drifting along general paths, enhancing a relatively simple nominal model using GP and solving the highly nonlinear constrained optimal control problem efficiently through an ADMM-based iLQR. The main contributions of our work are as follows:
\begin{enumerate}[label=\Roman*. ]
\item A hierarchical control framework is proposed for autonomous drifting, enabling vehicles to follow general paths. GPs are employed to correct model mismatches, significantly enhancing performance.
\item ADMM decomposes the optimization into two sub-problems, solved efficiently with iLQR and Quadratic Programming (QP), reducing computational burden.
\item Simulation results highlight the framework’s effectiveness, showing a 38$\%$ reduction in lateral error with the integration of GP and a 75$\%$ decrease in average computation time compared to IPOPT. Additionally, the method demonstrates robustness under varying friction conditions.

\end{enumerate}

\section{PRELIMINARIES}
\subsection{Gaussian Process Regression} 
GP regression is briefly introduced in the following. More details are available in \cite{williams2006gaussian}. A GP can be used as a nonparametric regression model to approximate a nonlinear function $g(z): \mathbb{R}^{n_z}\rightarrow\mathbb{R}$, using noisy observations y:
\begin{align}
\label{eq:gaussian noisy function}
    y=g(z)+\omega
\end{align}
where $z\in\mathbb{R}^{n_z}$ is the argument of unknown function $g(\cdot)$, $y\in \mathbb{R}$ is the noisy observation and $\omega \sim \mathcal{N}(0,\sigma^{2}_{\omega})$ is  the noise term. Consider N noisy function observations, denoted by $\textbf Z=[z_1,z_2,...z_N]\in \mathbb{R}^{n_z\times N}$ and $\textbf Y=[y_1,y_2,...y_N]^{T}\in \mathbb{R}^{N}$. The posterior distribution at a test point $z^*$  follows a Gaussian distribution with a mean and variance given by:
\begin{align}
\mu(z^*)& =K_{z^*\textbf Z} (K_{\textbf Z\textbf Z}+\textbf{I}\sigma^{2}_{\omega})^{-1}\textbf Y \\
\Sigma(z^*)&=K_{z^*z^*}-K_{z^*\textbf Z} (K_{\textbf Z \textbf Z}+\textbf{I}\sigma^{2}_{\omega})^{-1}K_{\textbf Z z^*}
\end{align}
where \textbf{I} is the identical matrix. $K_{(\cdot, \cdot)}$ represents the Gram matrix, $K_{\textbf Z z^*}={[k(z_1,z^*),k(z_2,z^*),...,k(z_N,z^*)]}^T\in\mathbb{R}^{N}$, $K_{z^*\textbf Z}{=(K_{\textbf Z z^*}})^T$, $K_{z^* z^*}=k(z^*,z^*)\in\mathbb{R}$ and $K_{\textbf Z\textbf Z}\in\mathbb{R}^{N \times N}$ with $[K_{\textbf Z\textbf Z}](i,j)=k(z_i,z_j)$. Here, $k(\cdot,\cdot)$ is the squared exponential (SE) kernel function
\begin{align}
k(z,z')={\sigma}_f^2  \text{exp}(-\frac{1}{2} (z-z')^\top\Lambda^{-1}(z-z'))
\end{align}
where $\Lambda \in \mathbb{R}^{N \times N}$ is a diagonal scaling matrix and ${\sigma_f}^2 \in \mathbb{R}$ is the covariance magnitude. Then, maximum likelihood estimation is applied to infer the unknown hyper-parameters:
\begin{align}
\max\limits_{{\sigma_f}^2,{\sigma_w}^2,\Lambda}\;\; -\frac{1}{2} \text{log}\left| K_{\textbf Z \textbf Z}\right|-\frac{1}{2} {\textbf Y}^T {K_{\textbf Z \textbf Z}} {\textbf Y}
\end{align}

For multi-dimension cases, each dimension is considered to be independently distributed and is trained separately, denoted by
\begin{align}
\label{eq:multi-dim GP}
g_{d}(z^*)\sim \mathcal{N}(g^{\mu}_{d}(z^*),g^\Sigma_{d} (z^*))
\end{align}
with $g^{\mu}_{d}= [\mu_{1}, \ldots, \mu_{n_z}]^\top \in\mathbb{R}^{N_z}$ and $g^\Sigma_{d} = \text{diag}([\sigma^2_{1}, \ldots, \sigma^2_{n_z}]) \in\mathbb{R}^{N_z \times N_z}$. 

\subsection{Iterative Linear Quadratic Regulator (iLQR)} 
Consider a nonlinear dynamic system $x_{i+1}=f(x_i,u_i)$ and the unconstrained optimal control problem (OCP) over N timesteps:
\begin{subequations}
\label{probabilistic problem}
\begin{align} 
\min\limits_{u_{1:N}}\;\;&l_f(x_{N+1}) +\sum^{N}_{i=1}l(x_i,u_i)\\
\;\text{s.t.}\;\;&x_{i+1}=f(x_{i},u_{i})
\end{align}
\end{subequations}

Instead of minimizing  over a sequence of control actions $u_{1:N}$, iLQR reduces it to minimization over every single action according to the Bellman equation:
\begin{align} \notag
V_i(x_i)&=\min_{u_i}Q_i(x_i,u_i)\\ 
&=\min_{u_i}l(x_i,u_i)+V_{i+1}(f(x_i,u_i))
\end{align}
where $Q_i$ denotes the state-action value function and $V_*$ denotes the value function of the state. Perturbed Q function $\delta Q_i$ is represented by:
\begin{align} \notag  
&\delta Q_i(\delta x_i,\delta u_i) =  Q_i(x_i+\delta x_i,u_i+\delta u_i)- Q_i(x_i, u_i) \\ 
&\approx \frac{1}{2}
\begin{bmatrix}
1 \\ \delta x_i \\ \delta u_i
\end{bmatrix}^T
\begin{bmatrix}
0       & (Q_i^T)_x & (Q_i^T)_u \\
(Q_i)_x & (Q_i)_{xx} & (Q_i)_{xu} \\
(Q_i)_u & (Q_i)_{ux} & (Q_i)_{uu}
\end{bmatrix}
\begin{bmatrix}
1 \\ \delta x_i \\ \delta u_i
\end{bmatrix} \label{eq:Perturbed Q function}
\end{align}
by performing Taylor expansion, with
\begin{align}\notag
(Q_i)_{x}&=(l_i)_{x}+{f_x^T}\cdot{(V_{i+1})_{x}} \\ \notag
(Q_i)_{u}&=(l_i)_{u}+{f_u^T}\cdot{(V_{i+1})_{x}} \\ \notag
(Q_i)_{xx}&=(l_i)_{xx}+{f_x^T}\cdot{(V_{i+1})_{xx}}\cdot{f_x} \\ \notag
(Q_i)_{ux}&=(l_i)_{ux}+{f_u^T}\cdot{(V_{i+1})_{xx}}\cdot{f_x} \\
(Q_i)_{uu}&=(l_i)_{uu}+{f_u^T}\cdot{(V_{i+1})_{xx}}\cdot{f_u}
\end{align}

By minimizing (\ref{eq:Perturbed Q function}), the optimal perturbed control action $\delta u_i^*$ for the perturbed Q-function is derived:
\begin{subequations}\label{eq:optimal perturbed control action}
\begin{align} 
\delta u_i^*&=k_i+K_i \delta x_i \\
k_i&=-(Q_i)_{uu}^{-1} (Q_i)_u\\
K_i&=-(Q_i)_{uu}^{-1} (Q_i)_{ux}
\end{align}
\end{subequations}

Substituting (\ref{eq:optimal perturbed control action}) into (\ref{eq:Perturbed Q function}) yields:
\begin{subequations} \label{eq:backward pass}
\begin{align}  
(V_i)_x&=(Q_i)_x-k_i^T {(Q_i)_{uu}} k_i \\
(V_i)_{xx}&=(Q_i)_{xx}-K_i^T {(Q_i)_{uu}} K_i
\end{align}
\end{subequations}

The above backward pass is conducted recursively until the first timestep. With a feasible nominal trajectory $\{{x_i},{u_i}\}$, we perform a forward pass to have a new feasible trajectory $\{{x^{+}_i},{u^+_i}\}$ :
\begin{align}  
{u^+_i}&={u_i}+\alpha k_i+K_i ({x^+_i}-x_i) \\
{x^+_{i+1}}&=f({x^+_i},{u^+_i})
\end{align}
where $0<\alpha\leq1$ is a backtracking line search parameter \cite{tassa2012synthesis}. We iterate between the forward pass and the backward pass until the objective function converges.

\section{Drift Dynamics with Residual Learning}
This section focuses on the dynamics of autonomous drifting, illustrating how GP is used to learn the model residuals. Additionally, an equilibrium analysis is performed to identify drifting equilibrium points.
\subsection{Nominal Vehicle Model}
\begin{figure}
    \centering
    \includegraphics[width=1\columnwidth]{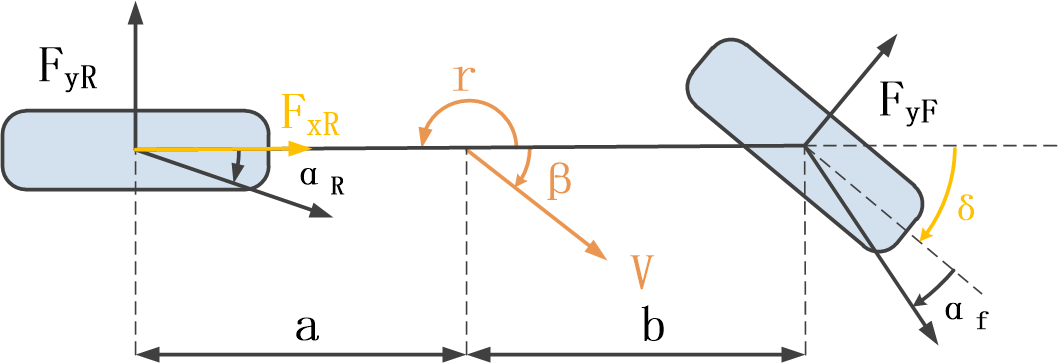}
    \caption{Single-track bicycle model for drifting  \cite{goh2020toward}.}
    \label{fig: bicycle model figure}
\end{figure}

Following \cite{goh2020toward}, a classic drift single-track bicycle model is applied, as illustrated in Fig. \ref{fig: bicycle model figure}. Simplifying the drifting vehicle as a rigid body, the model encompasses three states $x=[V,\beta,r]^T$: the total velocity vector with magnitude V, sideslip angle ${\beta}$ and yaw rate of the vehicle’s body r. The control variables $u=[\delta,{F}_{xr}]^T$ are steering angle $\delta$ and rear longitudinal force ${F}_{xr}$. The equations of motion then are:
\begin{align}
    \dot V &=\dfrac{-F_{yf} \text{sin}(\delta-\beta)+F_{yr} \text{sin}(\beta)+F_{xr} \text{cos}(\beta)}{m}  \label{eq:dot V} \\
    \dot \beta &=\dfrac{F_{yf} \text{cos}(\delta-\beta)+F_{yr} \text{cos}(\beta)-F_{xr} \text{sin}(\beta)}{mV}-r   \label{eq:dot beta}  \\
    \dot r &=\dfrac{aF_{yf} \text{cos}(\delta)-bF_{yr} }{I_{z}}  \label{eq:dot r}
\end{align}
where m represents the mass, ${I_{z}}$ is the moment of inertia in the vertical direction, a and b are the distance from the center of gravity to the front and rear axle respectively. The lateral forces acting on the vehicle are the front lateral force ${F}_{yf}$ and the rear lateral force ${F}_{yr}$, which are both modeled using the simplified Pacejka tire model \cite{bakker1987tyre}
\begin{align}
    F_{yf}=-\mu_f F_{zf} \text{sin}(C \text{arctan}(B\alpha_f))\\
    F_{yr}=-\mu_f F_{zr} \text{sin}(C \text{arctan}(B\alpha_r))
\end{align}
where $\mu_f$ is the friction coefficient, B and C are tire parameters, $F_{z\{f|r\}}$ is the vertical load on the tire. The front and rear tire slip angles are
\begin{align}
  \label{sideslip angle}
    \alpha_{f} &= \text{arctan}(\dfrac{V \text{sin}(\beta) + ar}{V\text{cos}(\beta)}) - \delta \\
    \alpha_{r} &= \text{arctan}(\dfrac{V \text{sin}(\beta) - br}{V \text{cos}(\beta)})
\end{align}

For the discrete-time nominal model formulation, the continuous system (\ref{eq:dot V}) - (\ref{eq:dot r}) is discretized with a first-order Euler discretization using a sampling time of $T_s$ = 100 ms. The discrete-time nominal model is denoted by
\begin{align}
\label{nominal model}
x_{i+1}=f_n(x_{i},u_{i})
\end{align}
\subsection{Residual Model Learning}  \label{subsection:residual model learning}
While the model above can reflect the vehicle dynamics to a considerable extent, there are still problems of model mismatch due to environmental interference or simplified formulas. To improve the model's accuracy and performance, we compensate the model error from the deviation of the nominal model by utilizing GP.

Throughout the vehicle's operation, performance data are systematically gathered $\{x_{1},u_{1},x_{2},u_{2},...\}$. GP's training inputs and outputs are
\begin{subequations}
\begin{align}
    z_i&=
    \begin{bmatrix}
    x_i \\ u_i
    \end{bmatrix}\\
    y_i&=x_{i+1}-f_n(x_i,u_i)
\end{align}
\end{subequations}

Due to the assumption that each dimension is uncorrelated, three independent GPs are trained (\ref{eq:multi-dim GP}). The stochastic vehicle model, incorporating residual compensation $g_{d}(x_i,u_i)$, is expressed as follows:
\begin{align}
\label{augmented model}
    x_{i+1}&=f_n(x_i,u_i)+g_{d}(x_i,u_i)
\end{align}

The performance of GP regression is highly dependent on the choice of data points. But retaining all data throughout the process will lead to computational infeasibility. So we only store the most informative data points in a dictionary \cite{hewing2019cautious}, maintaining the dictionary size based on points distance measurement method \cite{kabzan2019learning}.

\subsection{Drift Equilibrium Analysis} \label{subsection : Equilibrium Analysis}
Previous works have proved the existence of a drift saddle equilibrium point \cite{voser2010analysis} \cite{hindiyeh2014controller}. Drift equilibrium is a form of steady-state turning, which satisfies the formula for calculating the steady-state radius $R=V/r$. It's a common way to find the equilibrium point by simply setting vehicle states unchanged over control variables (i.e. $ x^{eq}=f_n(x^{eq},u^{eq})$) and fixing the values for {$R^{eq}$, $\delta^{eq}$}, making the number of unknowns equal to the number of equations. For the stochastic case (\ref{augmented model}), we use the predictive mean to describe the residual part (i.e. $x^{eq}=f_n(x^{eq},u^{eq})+g_d^{\mu}(x^{eq},u^{eq})$).

\begin{figure}
    \centering
    \includegraphics[width=1\columnwidth]{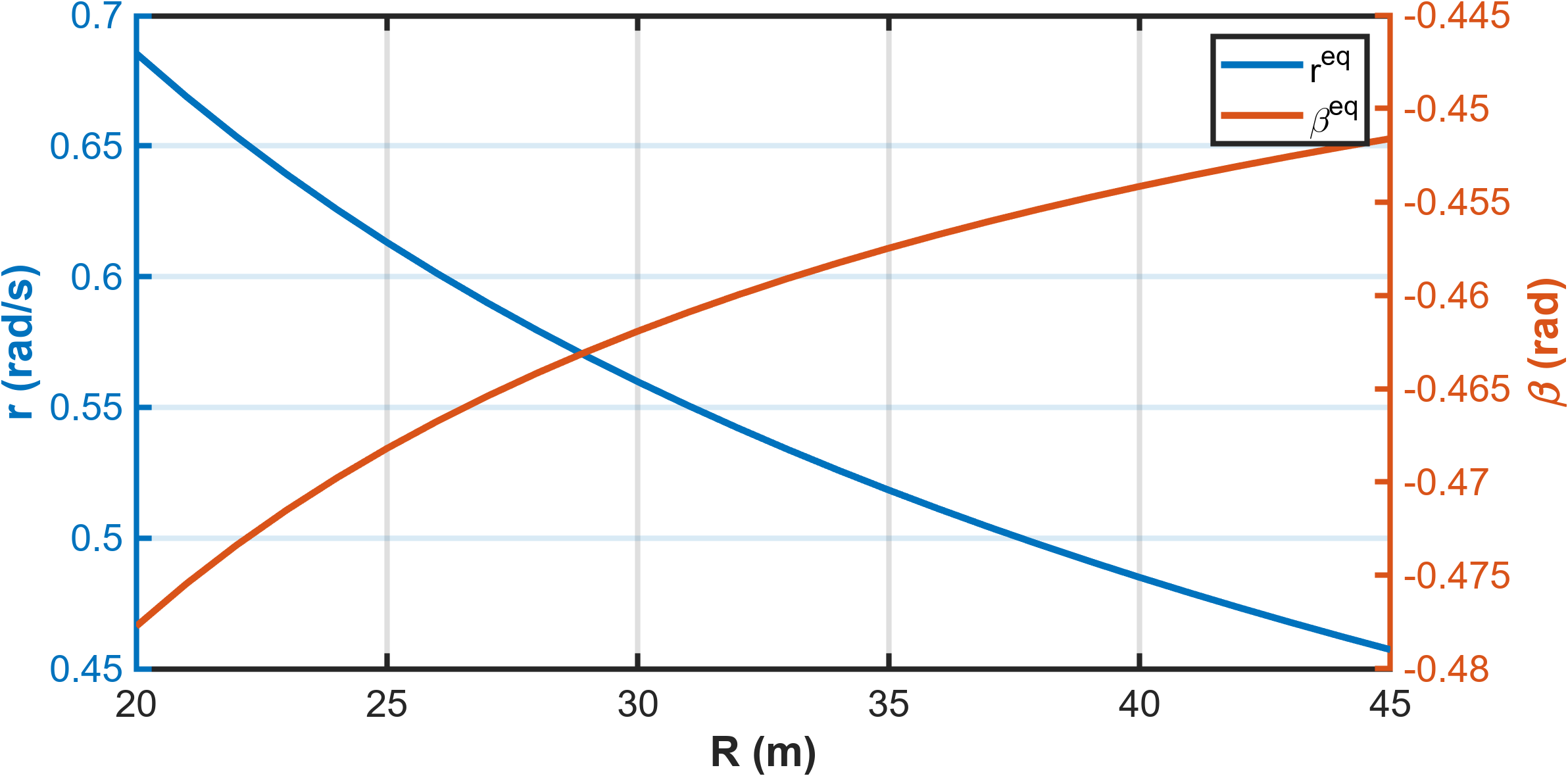}
    \caption{Equilibrium points for $\beta$ and $r$ with fixed $\delta^{eq}=-20$ deg and $R^{eq}$ varying from 20 m to 45 m.}
    \label{fig:equibrium figure}
\end{figure}

To get the equilibrium points, vehicle parameters are as follows: ${\mu}_f$=1, m=1140 kg, a=1.165 m, b=1.165 m, $I_z$=1020 kg$\cdot m^2$, B=12.55, C=1.494. Fig \ref{fig:equibrium figure} presents a portion of the equilibrium points derived from the nominal model (\ref{nominal model}) with $\delta^{eq}$=-20 deg and $R^{eq}$ varying from 20 m to 45 m.

\section{Drift Controller Structure}

 Our work focuses on steady-state drifting, aiming to maneuver the vehicle along a specific path while keeping the vehicle's states around drift equilibrium. An illustration of the framework can be found in Fig. \ref{fig:Overall system architecture}. The path tracking block is designed to provide equilibrium points for the optimal control block to follow. The optimal control block formulates a deterministic optimization problem and solves it with iLQR combined with ADMM. Throughout the process, data is collected for residual model learning with GP training (Section \ref{subsection:residual model learning}).

\subsection{Path Tracking}
\begin{figure}
    \centering
    \includegraphics[width=1\columnwidth]{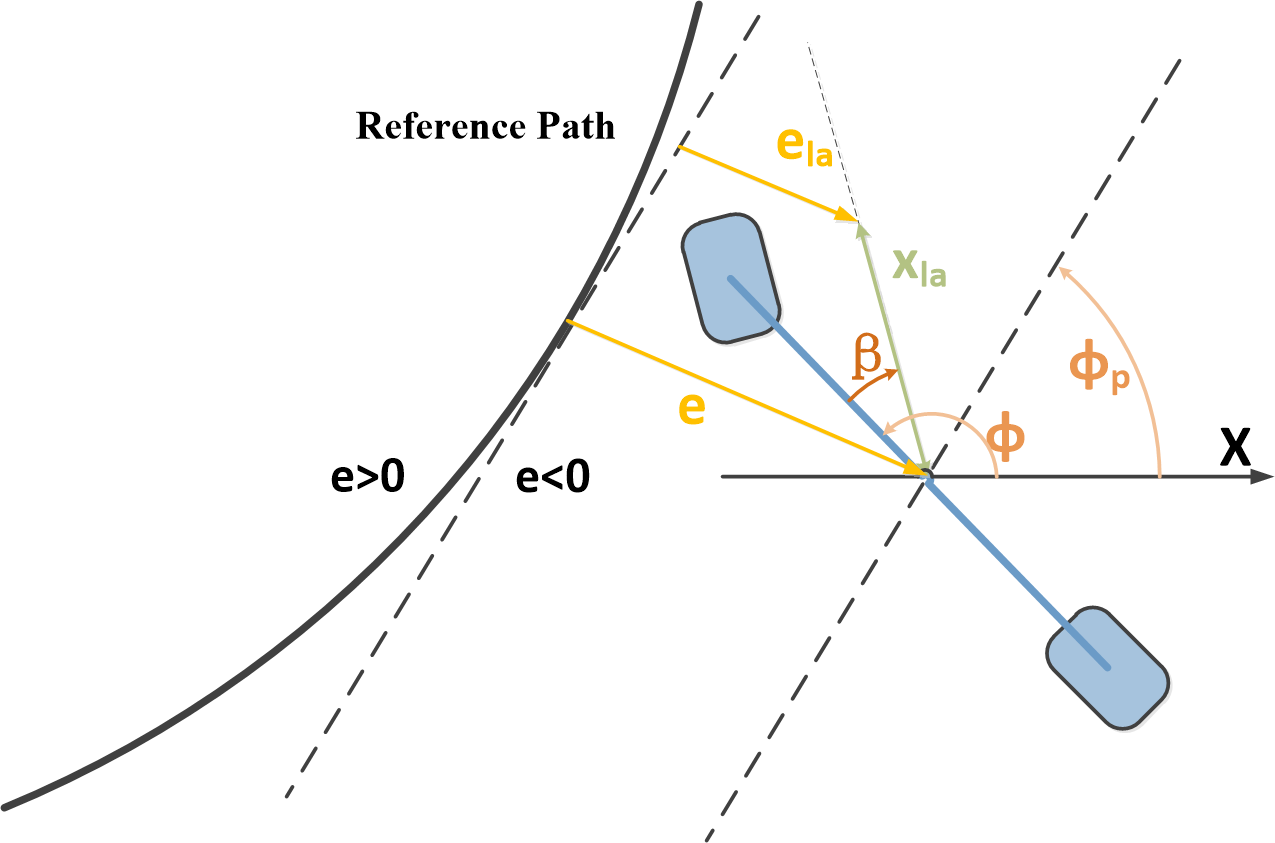}
    \caption{Error prediction with look-ahead distance.}
    \label{fig:LookAheadError figure}
\end{figure}

The core idea is to utilize the lateral error to adjust the desired curvature, as illustrated in Fig.\ref{fig:LookAheadError figure}. Instead of using the instantaneous lateral error $e$, we use the look-ahead error $e_{la}=e+x_{la}\text{sin}{(\Delta \phi)}$  as the input to a PID controller. Here, $x_{la}$ represents the look-ahead distance, and $\Delta \phi = \phi + \beta - \phi_{p} $ denotes the course direction error, where $\phi$ is the vehicle's heading angle and $\phi_{p}$ is the path angle. The PID controller’s output, which provides curvature compensation $\Delta k$, is used to determine the reference curvature $k^{eq} = \Delta k + k_{p}$, where $k_{p}$ is the path curvature. With $e_{la}$ accounting for potential trajectory deviations and the PID controller converting the error into the appropriate curvature adjustments, we can achieve a balance between fast response and strong robustness against minor disturbances. The desired drift equilibrium is then achieved with the specified $k^{eq}$ and a manually fixed $\delta^{eq}$ (Section \ref{subsection : Equilibrium Analysis})

\subsection{Optimal Control}
This part aims at following desired equilibrium points by solving optimal control problems.
\subsubsection{\textbf{Problem Formulation}}
Consider the following stochastic problem with control limitation and control-smoothing terms:
\begin{subequations}
\label{probabilistic problem}
\begin{align} 
\min\limits_{u_{1:N}}\;\;&\mathbb{E}[l_f(x_{N+1}) +\sum^{N}_{i=1}l(x_i,u_i)]+\sum^{N-1}_{i=1}\Vert{u_{i+1}-u_{i}}\Vert^2_{P}\\
\;\text{s.t.}\;\;&x_{i+1}=f_n(x_{i},u_{i})+g_d(x_{i},u_{i}) \\
&\underline{u}\leq u_{i} \leq \overline{u}
\end{align}
\end{subequations}
where the cost functions are in quadratic form: $l(x_i,u_i)=\Vert{x_i-x_\text{ref}}\Vert^2_Q+\Vert{u_i-u_\text{ref}}\Vert^2_R$ and $l_f(x_{N+1})=\Vert{x_i-x_\text{ref}}\Vert^2_{Q_f}$. $\underline{u}$ and $\overline{u}$ are lower and upper limits for control variables $u_{1:N}=[u_1^T,u_2^T \cdots u_{N}^T]^T \in \mathbb{R}^{n_u(N)}$.

Note that the learned GP model (\ref{augmented model}) has rendered states $x$ stochastic variables and states in fact are not Gaussian distributions because of the nonlinear mapping. A common approach to assess the uncertainty in these states is to approximate them as Gaussian distributions $x_i \sim \mathcal{N}(\mu_{i}, \Sigma_{i}) $ and propagate their mean and variance through successive linearizations \cite{kabzan2019learning}. In our experiment, we use the GP to estimate the state variances but do not propagate them as in an extended Kalman filter, in order to reduce the computational burden. The mean $\mu_{i}$ and variance $\Sigma_i$ are estimated as follows:
\begin{subequations}\label{eq:augmented states transfer}
\begin{align}   
&\mu_{i+1}=f_n(\mu_{i},u_{i})+g^{\mu}_ d(\mu_{i},u_{i}) \\
&\Sigma_{i+1}=\Sigma_{i}+g^{\Sigma}_d(\mu_{i},u_{i}) 
\end{align}
\end{subequations}

We reformulate the probabilistic problem (\ref{probabilistic problem}) to a deterministic one by introducing the Gaussian belief augmented state vector \cite{pan2014probabilistic} $\overline{x}_i=[\mu_i^T, vec(\Sigma_i)^T]^T$:
\begin{subequations}
\label{deterministic problem}
\begin{align} 
\min\limits_{u_{1:N}}\;\;&L_f(\overline{x}_{N+1}) +\sum^{N}_{i=1}L(\overline{x}_i,u_i)+\sum^{N-1}_{i=1}\Vert{u_{i+1}-u_{i}}\Vert^2_{P}\\
\;\text{s.t.}\;\;&{\overline{x}}_{i+1}=F(\overline{x}_{i},u_i)\\
&\underline{u}\leq u_{i} \leq \overline{u}
\end{align}
\end{subequations}
where the new cost function is defined as $L(\overline{x}_i,u_i)=\Vert{\mu_i-x_\text{ref}}\Vert^2_Q+\text{tr}(Q\Sigma_i)+\Vert{u_i-u_\text{ref}}\Vert^2_R$ and the augmented dynamics are given by $F(\overline{x}_{i},u_i)$ (\ref{eq:augmented states transfer}). 

It is important to highlight that introducing GP can significantly increase the nonlinearity of the system. This makes the choice of a highly efficient solver critical to ensuring timely and accurate solutions. To address this, we solve the deterministic problem (\ref{deterministic problem}) using a numerical solver that integrates iLQR with ADMM, which notably reduces the average computation time. This approach shares a fundamental alignment with the methodology presented in \cite{ma2022alternating}.

\subsubsection{\textbf{ADMM-based iLQR}}
In this part, ADMM is employed to exploit the structure of (\ref{deterministic problem}) by decomposing the optimization problem into two manageable sub-problems: a typical unconstrained OCP and a QP problem. Firstly, we define the indicator function:
\begin{align} 
I_{\mathcal{A}}(a) = 
\begin{cases}
+\infty & \text{if } a \notin \mathcal{A},  \\
    0 & \text{else}.
\end{cases}
\end{align}
with respect to a set $\mathcal{A}$. We further define sets:
\begin{align} 
{\mathcal{B}}&=\{   b=[\overline{x}_1\dots \overline{x}_{N+1};w_1\dots w_{N}] |{\overline{x}}_{i+1}=F(\overline{x}_{i},w_i)\}\\
{\mathcal{C}}&=\{ c=[u_1;u_2\dots u_{N}] \;|\; \underline{u}\leq u_{i} \leq \overline{u}\}
\end{align}

By introducing the consensus variable $w_{1:N}$, (\ref{deterministic problem}) can be transferred as:
\begin{subequations}
\label{ADMM problem-1}
\begin{align} 
\min\limits_{\underset{u_{1:N}}{w_{1:N}}}\;\;&L_f(\overline{x}_{N+1}) +\sum^{N}_{i=1}L(\overline{x}_i,w_i) +I_\mathcal{B}([\overline{x}_{1:N+1};w_{1:N}])\\ 
&+\sum^{N-1}_{i=1}\Vert{u_{i+1}-u_{i}}\Vert^2_{P}+I_\mathcal{C}(u_{1:N})\\
\;\text{s.t.}\;\;&w_{1:N}=u_{1:N}
\end{align}
\end{subequations}

Then, we can define the augmented Lagrangian function of (\ref{ADMM problem-1}) as:
\begin{subequations}
\begin{align} 
\label{ADMM augmented Lagrangia problem}
\min\limits_{\underset{u_{1:N}}{w_{1:N}}}\;&J=L_f(\overline{x}_N) +\sum^{N}_{i=1}L(\overline{x}_i,w_i) +I_\mathcal{B}([\overline{x}_{1:N+1};w_{1:N}])\\
&+\sum^{N-1}_{i=1}\Vert{u_{i+1}-u_{i}}\Vert^2_{P}+I_\mathcal{C}(u_{1:N})\\
&+\sum^{N}_{i=1}{\lambda_i}^T(w_{i}-u_{i})+\frac{\rho}{2}\Vert{w_{i}-u_{i}}\Vert^2
\end{align}
\end{subequations}
where $\lambda_{1:N}=[\lambda_1^T,\lambda_2^T \cdots \lambda_{N}^T]^T \in \mathbb{R}^{n_u N}$ is a Lagrange multiplier and $\rho \in \mathbb{R}$ is a scalar penalty weight. We arrive at the three-step
ADMM iteration \cite{boyd2011distributed} by alternately minimizing over $w$ and $u$, rather than simultaneously minimizing over both:
\begin{align} 
w^{+}_{1:N}&=\arg\min\limits_{w}\;\;J(w_{1:N},u_{1:N},\lambda_{1:N})  \label{eq:w update} \\ 
u^{+}_{1:N}&=\arg\min\limits_{u}\;\;J(w^{+}_{1:N},u_{1:N},\lambda_{1:N})  \label{eq:u update} \\ 
\lambda^{+}_{1:N}&=\lambda_{1:N}+\rho \;(w^{+}_{1:N}-u^{+}_{1:N})\label{eq:lambda update}
\end{align}

The above three steps are repeated until the desired convergence tolerance is reached, with further details provided below.

{The first ADMM step} (\ref{eq:w update}) is equivalent to solving the following problem:
\begin{subequations}
\label{eq:first ADMM step}
\begin{align} 
\arg\min\limits_{w}\;\;&L_f(\overline{x}_{N+1}) +\sum^{N}_{i=1}\overline{L}(\overline{x}_i,w_i)\\
\;\text{s.t.}\;\;&{\overline{x}}_{i+1}=F(\overline{x}_{i},w_i)
\end{align}
\end{subequations}
with $\overline{L}(\overline{x}_i,w_i)=L(\overline{x}_i,w_i)+{\lambda_i}^T(w_{i}-u_{i})+\frac{\rho}{2}\Vert{w_{i}-u_{i}}\Vert^2$. Since the problem only involves dynamic constraints, the iLQR algorithm can efficiently solve the optimal control problem (\ref{eq:first ADMM step}) without requiring an initial feasible solution.

{The second ADMM step} (\ref{eq:u update}) is equivalent to solving the following QP problem:
\begin{subequations}
\label{eq:second ADMM step}
\begin{align} 
\arg\min\limits_{u}\;\;&\sum^{N-1}_{i=1}\Vert{u_{i+1}-u_{i}}\Vert^2_{P}\\ \notag
&+\sum^{N}_{i=1}[{\lambda_i}^T(w^+_{i}-u_{i})+\frac{\rho}{2}\Vert{w^+_{i}-u_{i}}\Vert^2]
\;\\
\text{s.t.}\;\;&\underline{u}\leq u_{i} \leq \overline{u}
\end{align}
\end{subequations}

A novel fast solver for QP based on Pseudo-Transient Continuation (PTC) \cite{calogero2024enhanced} is applied for solving the above problem (\ref{eq:second ADMM step}).
\footnote{\cite{calogero2024enhanced} requires equality constraints with full row rank. To solve this problem, we introduce an additional constant variable (i.e. $[u_{1:N};c_0]$) to have an equality constraint (i.e. $c_0=0$). Please refer to the original paper for more details.}

{The third ADMM step} (\ref{eq:lambda update}) is a gradient-ascent update on the Lagrange multiplier.

\section{SIMULATION RESULTS}
The whole experiment is conducted on a joint simulation platform using MATLAB-R2023a and the high-fidelity vehicle simulation software CarSim-2019. The simulation is executed on an Intel i7-8565U processor.

\subsection{Simulation Setup}

We selected a Clothoid curve as the reference path \cite{goh2020toward}, with a radius ranging between 20 m and 45 m, as shown in Fig.\ref{fig:trajectory figure}. The front wheel angle $\delta^{eq}$ is set to -20 deg, with an initial speed of 12 m/s.  The look-ahead distance $x_{la}$ is 30 m. Vehicle parameters have been illustrated in section \ref{subsection : Equilibrium Analysis}. The real friction coefficient is set to $\mu_f$=1.

The drift controller is activated at the start of the simulation, operating independently without the help of a guide controller. In all scenarios, the prediction horizon is chosen as N = 20. The weighting parameters are established as $Q=Q_f=\text{diag}(0.1, 1, 1)$, $R=\text{diag}(1,10^{-7})$ and $P=\text{diag}(10,10^{-7})$. After successfully solving the optimal control problem, the results are stored and then used as the initial guess for the next timestep to facilitate warm starting.

The simulation is initialized with the nominal model, where all GP-dependent variables are initially set to zero (\ref{augmented model}). The data collected under the nominal controller is then used to populate the GP dictionary with an initial set of up to 50 data points for each dimension. After completing the first lap, the GP-enhanced dynamics (\ref{eq:augmented states transfer}) are employed for the subsequent 5 laps.

\subsection{Results}
\begin{figure}
    \centering
    \includegraphics[width=0.95\columnwidth]{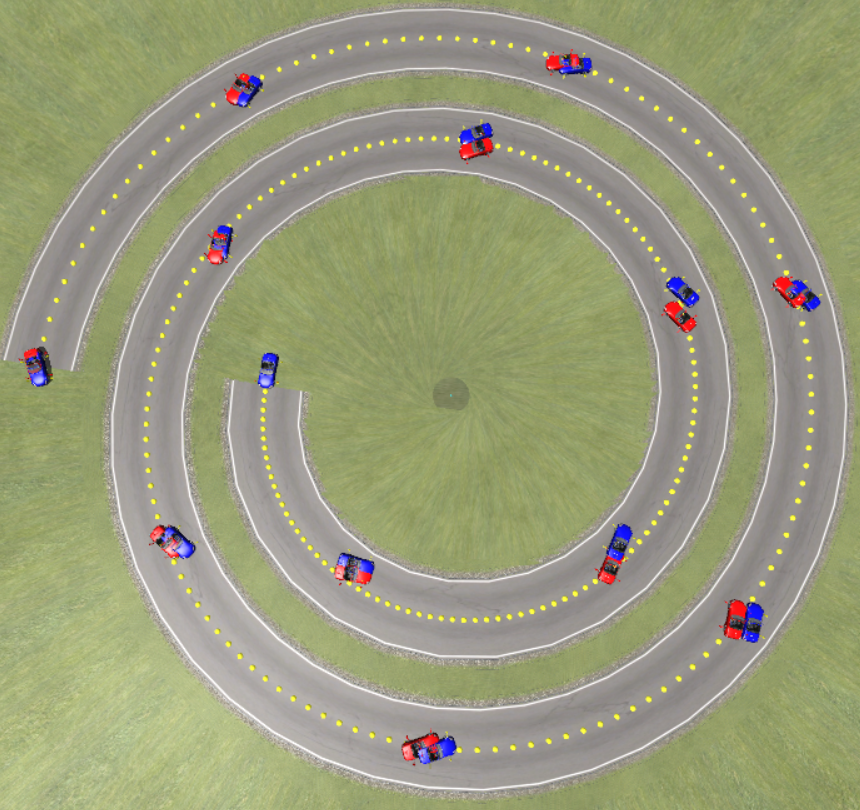}
    \caption{The vehicle trajectories for the nominal controller (in blue) and the GP-based controller in the 6th lap (in red) are shown, with yellow points indicating the reference path.}
    \label{fig:trajectory figure}
\end{figure}

\begin{figure}
    \centering
    \includegraphics[width=1\columnwidth]{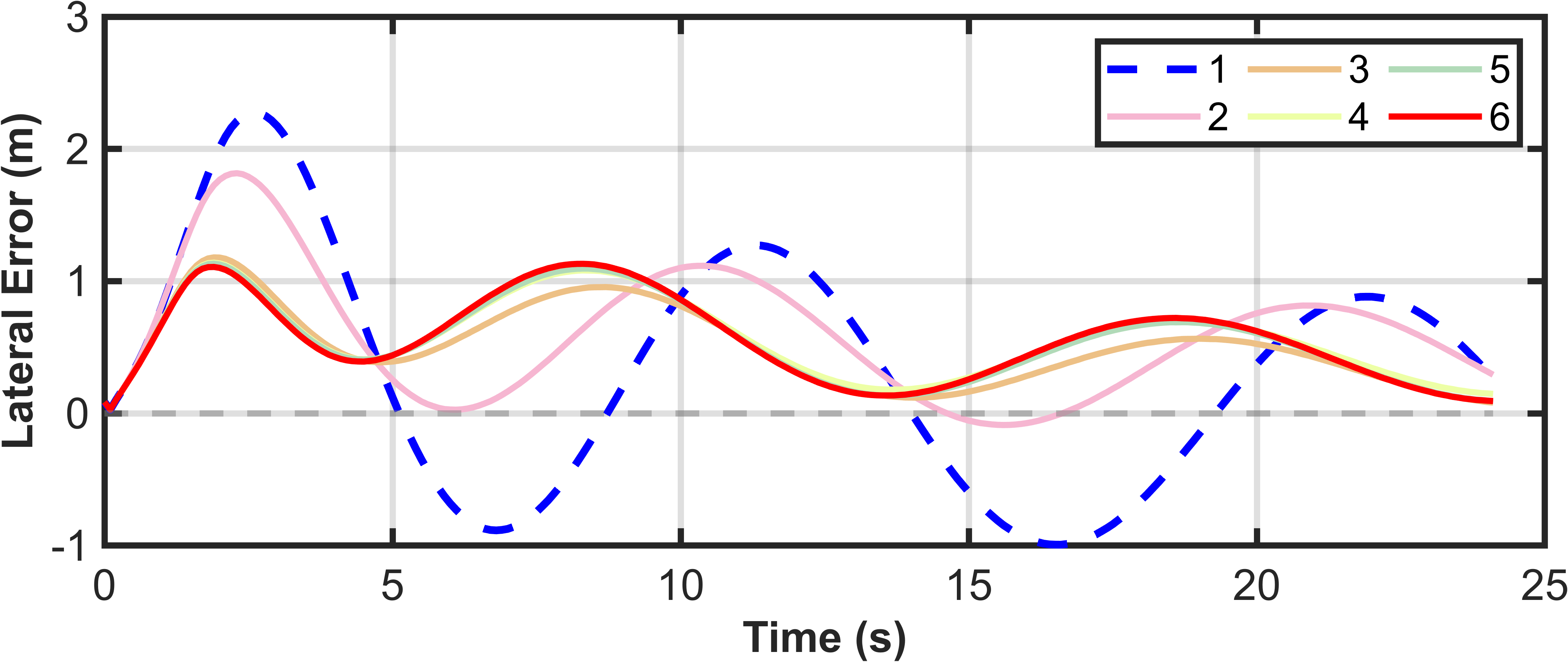}
    \caption{Lateral error across all 6 laps. Average tracking error gradually decreases as the learning process progresses.}
    \label{fig:lateralError figure}
\end{figure}

\begin{figure}
    \centering
    \includegraphics[width=1\columnwidth]{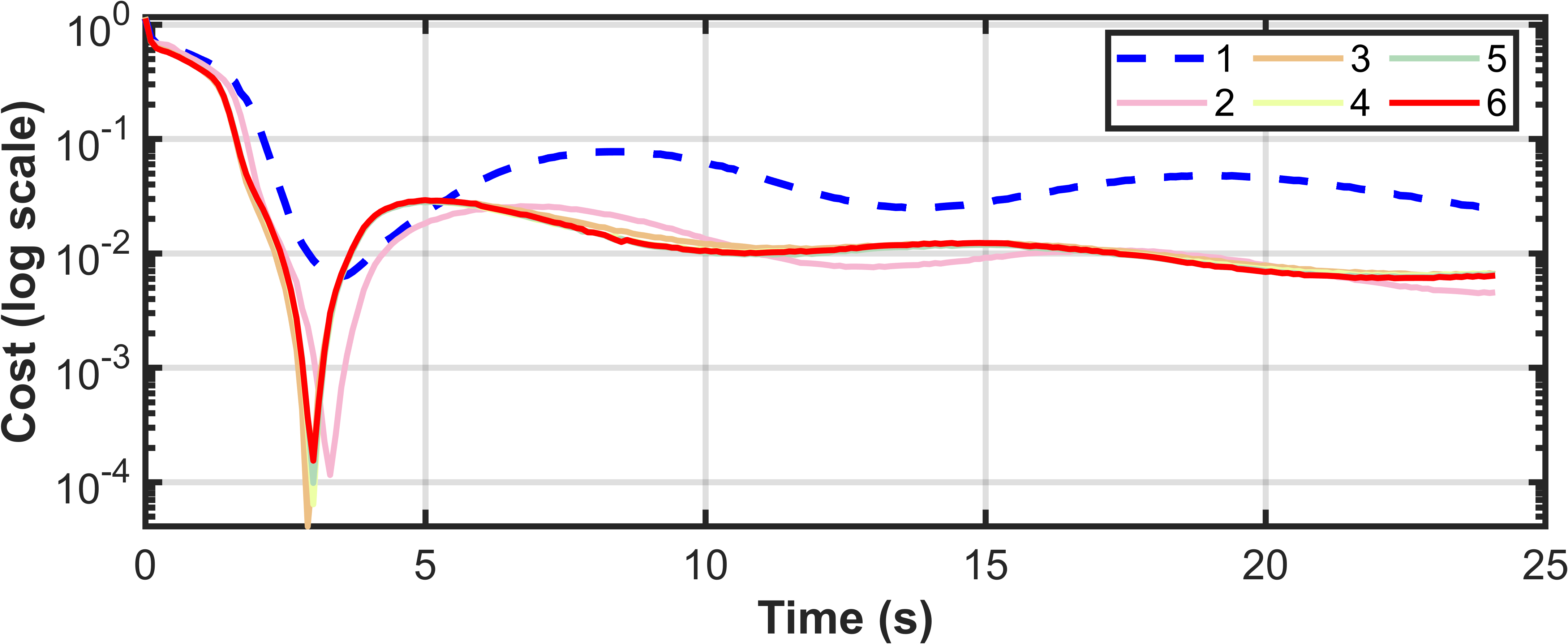}
    \caption{Equilibrium point tracking performance over six laps, with GP-enhanced controllers exhibiting lower tracking costs.}
    \label{fig:cost figure}
\end{figure}

To demonstrate the trajectory tracking effectiveness of the drifting control approach, we first compare the lateral errors across successive laps, as shown in Fig.\ref{fig:lateralError figure} and summarized in Table.\ref{tab:performance table}. Initially, with the nominal controller, the Root Mean Square Error (RMSE) is approximately 0.94 m, and the maximum lateral error is around 2.28 m. A significant improvement is observed in lap 3, with the RMSE and maximum lateral error reduced to 0.58 m and 1.18 m respectively, constituting an improvement of almost $38\%$ and $48\%$. The drifting trajectories for the nominal controller (lap 1) and the GP-based controller (lap 6) are illustrated in Fig \ref{fig:trajectory figure}.

In addition, we examine the equilibrium tracking performance. It is important to note that the GP is also utilized in the drift equilibrium calculation to provide more accurate desired states. To assess how well the system tracks the changing equilibrium point, we employ a quadratic cost function $\text{Cost}(i)=\Vert{x(i)-x_\text{ref}(i)}\Vert^2_Q+\Vert{u(i)-u_\text{ref}(i)}\Vert^2_R$. The equilibrium point tracking ability is shown in Fig \ref{fig:cost figure}. The GP-based controller exhibits superior equilibrium tracking performance throughout each lap, as evident by its lower cost compared to the controller without GP. And the average cost converges quickly in the subsequent laps. 
\begin{figure}
    \centering
    \includegraphics[width=1\columnwidth]{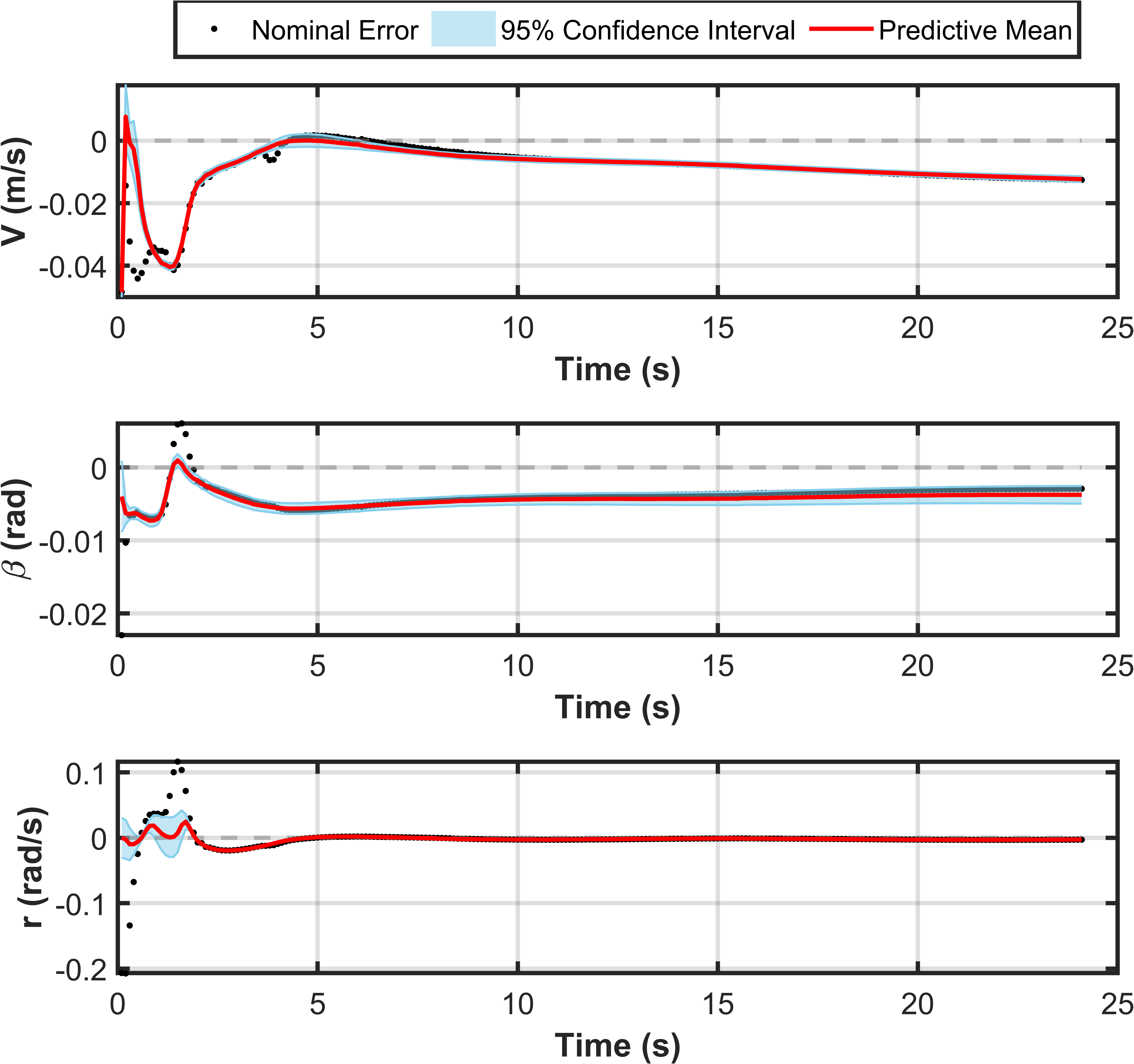}
    \caption{Nominal model error (black dots) and GP compensation (red line). Light blue shadows are $95\%$ confidence interval.}
    \label{fig:statesCompensation figure}
\end{figure}

Next, the model's learning performance is analyzed, as illustrated in Fig. \ref{fig:statesCompensation figure}. We present the nominal model errors encountered (i.e. errors between the measurement and nominal prediction), along with the corresponding GP compensations in the final lap. The results indicate that both the GP's mean and uncertainty estimates align well with the true model errors. To quantify the learning performance, we define the prediction error as $\Vert x_{i+1}-f_n(x_i,u_i)\Vert$ for the nominal controller, and $\Vert x_{i+1}-f_n(x_{i},u_{i})-g^{\mu}_d(x_{i},u_{i})\Vert$ for the GP-based controller. As shown in Table \ref{tab:performance table}, the prediction error decreased from 0.0153 at the start to 0.0059 after two rounds of learning, reflecting a reduction of approximately $61\%$.

\begin{figure}
    \centering
    \includegraphics[width=1\columnwidth]{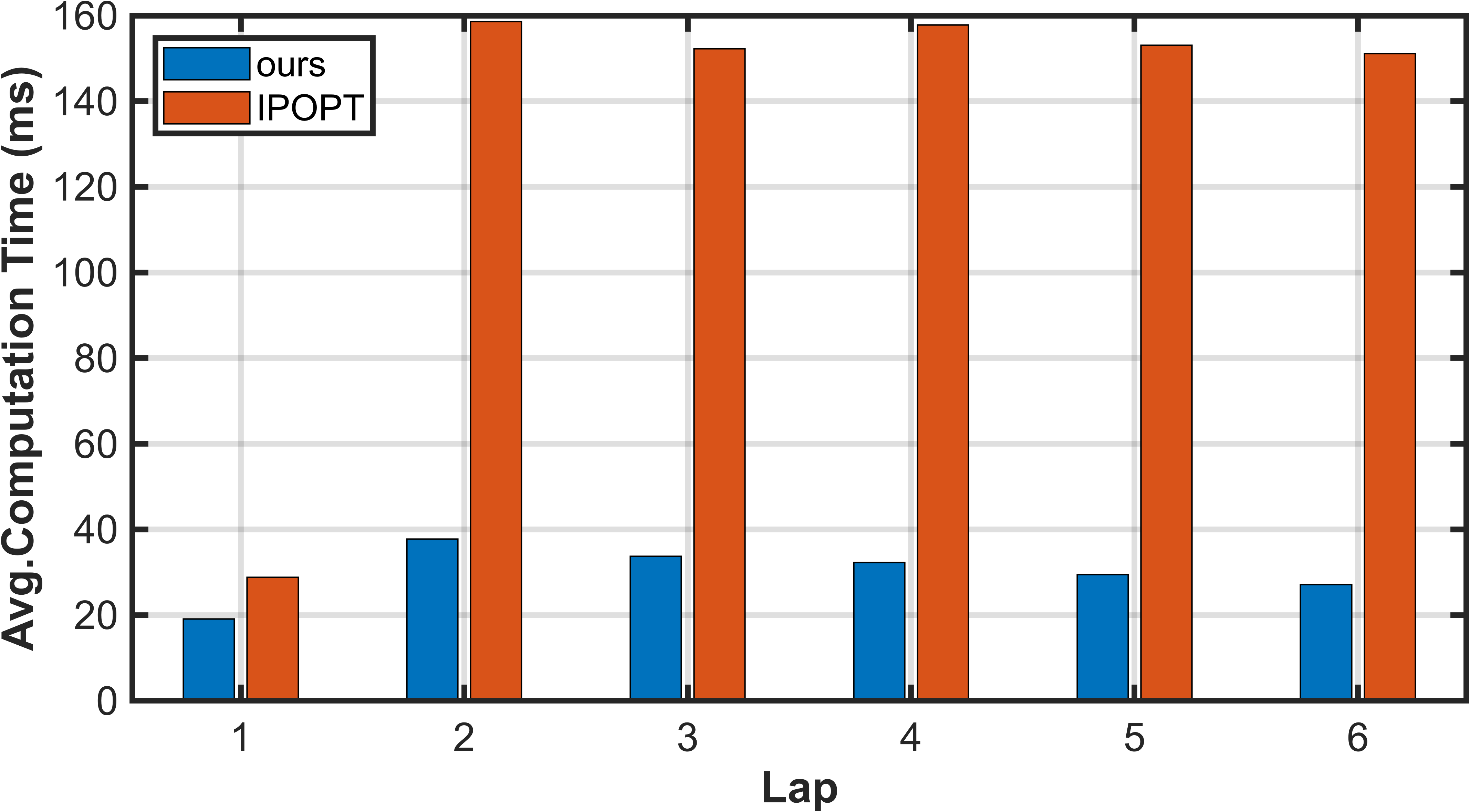}
    \caption{Computation time comparison performed on an Intel i7-8565U processor.}
    \label{fig:computationTime figure}
\end{figure}

The computation time of our ADMM-based iLQR method is recorded, and comparison studies are conducted using CasADi \cite{Andersson2019}, an optimization solver that utilizes IPOPT \cite{wachter2006implementation}. The results, shown in Fig. \ref{fig:computationTime figure}, indicate a significant increase in computation time for the IPOPT following the introduction of GP, consistently reaching around 160 ms from lap 2 onward. In contrast, our method demonstrates much lower computation time, remaining under 20 ms in lap 1 and below 40 ms in the subsequent laps. The time cost decreased about 75$\%$ from lap 2 onward. These findings highlight the strong potential of our approach for real-time applications and higher control frequency.

To rigorously evaluate the robustness of the proposed method, a 10$\%$ variation (both increasing and decreasing) in the friction coefficient $\mu_f$ was applied to simulate fluctuations in the vehicle's operating environment. The system's performance was systematically evaluated under these varying conditions using the established criteria. As indicated in Table \ref{tab:performance table}, the method consistently demonstrated high effectiveness, underscoring its robustness and adaptability across a range of dynamic conditions, thereby ensuring reliable operation.

\begin{table*}[htb] 
    \centering
    \caption{Comparison of drift performance under different friction coefficients}
    \setlength{\tabcolsep}{12pt} 
    \begin{tabular}{c c c c c c c} 
        \toprule
        \multirow{2}{*}{\textbf{Metric}} & \multirow{2}{*}{\textbf{Lap}} & \multicolumn{5}{c}{\textbf{Friction Coefficient}} \\
        \cmidrule(lr){3-7}
        & & \textbf{$\mu_f=1.10$} & \textbf{$\mu_f=1.05$} & \textbf{$\mu_f=1.00$} & \textbf{$\mu_f=0.95$} & \textbf{$\mu_f=0.90$} \\
        \midrule
        \multirow{6}{*}{RMSE. Lateral Error (m)} 
            & 1 & $\Delta$2.7652  & 1.6243 & 0.9405 & 0.8868 & 1.5080 \\
            \cdashline{2-7} 
            & 2 & 1.6414 & \textbf{0.6827} & 0.7612 & 0.5187 & 1.1663 \\
            & 3 & 0.9084 & 0.8064 & \textbf{0.5839} & 0.5823 & 1.1190 \\
            & 4 & \textbf{0.6265} & 0.9116 & 0.6412 & 0.6037 & \textbf{1.0927} \\
            & 5 & 0.7392 & 0.7507 & 0.6340 & \textbf{0.5112} & 1.1138 \\
            & 6 & 0.6499 & 0.7798 & 0.6436 & 0.5527 & 1.1232 \\
        \midrule
        \multirow{6}{*}{Max. Lateral Error (m)} 
            & 1 & $\Delta$5.3484  & 3.5088 & 2.2770 & 1.5935 & 2.7996 \\
            \cdashline{2-7} 
            & 2 & 3.2196 & 1.7688 & 1.8167 & 0.9979 & 1.9847 \\
            & 3 & 1.8731 & 1.9754 & 1.1799 & 1.1593 & 1.9624 \\
            & 4 & \textbf{1.3426} & 2.2271 & 1.1399 & 1.1968 & \textbf{1.9551} \\
            & 5 & 1.6108 & 1.6994 & 1.1319 & \textbf{0.9777} & 1.9714 \\
            & 6 & 1.5402 & \textbf{1.5899} & \textbf{1.1312} & 1.1302 & 1.9863 \\
        \midrule
        \multirow{6}{*}{Avg. Cost} 
            & 1 & $\Delta$0.5159  & 0.0999 & 0.0798 & 0.0770 & 0.0931 \\
            \cdashline{2-7} 
            & 2 & 0.1124 & 0.0594 & 0.0497 & \textbf{0.0471} & \textbf{0.0605} \\
            & 3 & 0.0663 & 0.0562 & 0.0456 & 0.0479 & 0.0626 \\
            & 4 & 0.0573 & 0.0570 & 0.0450 & 0.0483 & 0.0639 \\
            & 5 & 0.0543 & 0.0491 & \textbf{0.0448} & 0.0474 & 0.0645 \\
            & 6 & \textbf{0.0504} & \textbf{0.0474} & 0.0473 & 0.0503 & 0.0648 \\
        \midrule
        \multirow{6}{*}{Avg. Prediction Error} 
            & 1 & $\Delta$0.0428  & 0.0319 & 0.0153 & 0.0235 & 0.0471 \\
            \cdashline{2-7} 
            & 2 & 0.0420 & 0.0142 & 0.0065 & 0.0086 & 0.0308 \\
            & 3 & 0.0298 & 0.0060 & \textbf{0.0059} & 0.0083 & 0.0296 \\
            & 4 & 0.0168 & 0.0049 & 0.0060 & 0.0081 & \textbf{0.0295} \\
            & 5 & 0.0067 & \textbf{0.0049} & 0.0060 & \textbf{0.0072} & 0.0296 \\
            & 6 & \textbf{0.0067} & 0.0053 & 0.0060 & 0.0078 & 0.0296 \\
        \bottomrule
    \end{tabular}
    \label{tab:performance table}

    \begin{minipage}{\linewidth} 
        \vspace{1mm} 
        \footnotesize
        \textit{Note:} $\Delta$ denotes a failure to drift due to an excessively large $\mu_f$. However, with the integration of GP, the vehicle successfully learns to drift starting from lap 2.
    \end{minipage}
\end{table*}

\section{CONCLUSIONS}

This work presents a novel hierarchical control framework combining GP with an ADMM-based iLQR for autonomous vehicle drift control. The integration of GP effectively compensates for model inaccuracies, significantly improving path-tracking performance. The combination of ADMM and iLQR allows for efficient problem decomposition, leading to efficient solutions. Simulation results show that our method achieves a 38$\%$ reduction in RMSE lateral error and a 75$\%$ reduction in computation time compared to IPOPT, making it highly suitable for real-time applications in dynamic environments.

Future work will focus on extending the framework to experimental validation using scaled autonomous racing cars, exploring its adaptability to real-world scenarios, and further optimizing its performance under varying environmental conditions.

\addtolength{\textheight}{-0cm}   




\section*{ACKNOWLEDGMENT}
 Thanks to Bei Zhou and Xiewei Xiong for valuable suggestions.

\bibliography{references}
\bibliographystyle{IEEEtran}

\end{document}